\documentclass{article}
\usepackage[preprint]{colm2026_conference}

\usepackage{amsmath,amssymb,mathtools}

\usepackage{microtype}
\usepackage{booktabs}
\usepackage{multirow}
\usepackage{makecell}
\usepackage[bottom]{footmisc}
\usepackage{lineno}

\usepackage{graphicx}
\usepackage{tikz}
\usetikzlibrary{positioning,arrows.meta,decorations.pathmorphing}

\usepackage{algorithm}
\usepackage{algpseudocode}

\usepackage{tcolorbox}
\tcbuselibrary{skins}

\usepackage{hyperref,url}
\definecolor{darkblue}{rgb}{0,0,0.5}
\hypersetup{colorlinks=true,citecolor=darkblue,linkcolor=darkblue,urlcolor=darkblue}

\tikzset{
  nd/.style={rectangle,draw=gray!60,fill=white,rounded corners=2pt,
             minimum height=1.6em,text width=5em,align=center,font=\small},
  nd_b/.style={nd,draw=darkblue,fill=blue!7,text=darkblue,font=\small\bfseries},
  nd_o/.style={nd,draw=orange!80!black,fill=orange!8,
               text=orange!80!black,font=\small\bfseries},
  nd_g/.style={nd,draw=green!60!black,fill=green!6,
               text=green!60!black,font=\small\bfseries},
  arr/.style={->,thick,color=darkblue!50,>=Stealth},
  darr/.style={arr,dashed,color=darkblue!30},
}

\newcommand{\thinkreset}{\textsc{ThinkReset}}
\newcommand{\Rs}{R_{\mathrm{succ8}}}
\newcommand{\triggerratio}{\alpha}

\title{ThinkReset: Learnable Intermediate Interface Construction for Bounded-Context Long-Horizon Reasoning}

\author{
Fei Ding\textsuperscript{1}\thanks{Corresponding author: \texttt{dignfei@gmail.com}.}
\quad Yongkang Zhang\textsuperscript{1}
\quad Runhao Liu\textsuperscript{1}
\quad Yuhao Liao\textsuperscript{2}
\quad Zijian Zeng\textsuperscript{2}\\
\textsuperscript{1}Alibaba Group
\quad
\textsuperscript{2}Tsinghua University
}

\begin{document}
\ifcolmsubmission\linenumbers\fi
\maketitle

\begin{abstract}
Long chain-of-thought reasoning improves performance on complex problems, but it also introduces redundancy accumulation, context overflow, and error anchoring. We argue that under bounded context windows, the core bottleneck is not trajectory compression or test-time control, but the absence of a reusable intermediate interface that can replace discarded history and support continued solving. We further identify a key failure mode of outcome-reward-driven long-chain reinforcement learning: when the model has not solved the task before the window is nearly exhausted, the final-answer reward encourages premature guessing rather than continued careful reasoning. We propose \thinkreset{}, a text-space instantiation of this view. \thinkreset{} explicitly constructs reusable intermediate interfaces through interface writeback and reset, and directly optimizes post-reset continuation success. Across multiple long-horizon reasoning benchmarks, this perspective consistently improves success rates under fixed context windows.
\end{abstract}

\begin{figure}[!b]
    \centering
    \scalebox{0.88}{
        \begin{tikzpicture}[
            box/.style={rectangle, draw=gray!50, fill=white, rounded corners=2.5pt,
                minimum width=1.45cm, minimum height=0.72cm, align=center,
                font=\scriptsize},
            wrongbox/.style={box, draw=red!50, fill=red!6, text=red!70!black},
            goodbox/.style={box, draw=green!60!black, fill=green!6, text=green!60!black},
            summbox/.style={box, draw=blue!60!black, fill=blue!7, text=blue!60!black,
                minimum width=3.0cm, minimum height=0.82cm,
                font=\scriptsize\bfseries},
            arr/.style={->, thick, >=Stealth, color=gray!60},
            redarr/.style={arr, color=red!50},
            bluearr/.style={arr, color=blue!50},
            label/.style={font=\small\bfseries, align=center},
            sublabel/.style={font=\scriptsize, color=black!75, align=center},
            ]

            \node[label] at (0.0, 2.85) {(a) Trajectory-centric view};
            \node[sublabel] at (0.0, 2.55) {history keeps growing};

            \node[box]      (t1) at (-3.0, 1.75) {Step\\$t_1$};
            \node[wrongbox] (t2) at ( 0.0, 1.75) {Wrong\\hypothesis\\$t_2$};
            \node[box]      (t3) at ( 3.0, 1.75) {Step\\$t_3$};

            \draw[arr] (t1) -- (t2);
            \draw[redarr] (t2) -- (t3);
            \node[font=\tiny, color=red!60, fill=white, inner sep=1pt] at (1.45, 2.08) {Error anchoring};

            \draw[thick, dashed, color=red!60] (4.7, 2.15) -- (4.7, 1.05);
            \node[font=\tiny, color=red!60, align=center] at (5.85, 1.6) {Window exhausted\\forced truncation};

            \node[label] at (0.0, -0.18) {(b) Interface construction with \textsc{ThinkReset}};
            \node[sublabel] at (0.0, -0.48) {construct a reusable interface};

            \node[box]      (r1) at (-3.0, -1.28) {Step\\$t_1$};
            \node[wrongbox] (r2) at ( 0.0, -1.28) {Wrong\\hypothesis\\$t_2$};
            \node[box]      (r3) at ( 3.0, -1.28) {Step\\$t_3$};
            \node[goodbox]  (r4) at ( 5.2, -1.28) {Continue\\solving\\$t_4'$};

            \draw[arr] (r1) -- (r2);
            \draw[redarr] (r2) -- (r3);

            \node[summbox]
            (summ) at (0.0, -2.6) {Reusable intermediate\\interface\\for continued solving};
            \node[font=\scriptsize\bfseries, color=blue!70] at (0.0, -3.25) {\texttt{[RESET]}};

            \draw[bluearr] (r3.south) to[out=-70,in=20] (summ.east);
            \draw[bluearr] (summ.east) to[out=15,in=-120] (r4.south);

            \node[font=\tiny, color=red!50, fill=white, inner sep=1pt] at (0.0, -1.95) {replace original history};

    \end{tikzpicture}}
    \caption{Trajectory retention versus interface construction under bounded context windows. Instead of continuing to grow the autoregressive history, \thinkreset{} replaces it with a reusable intermediate interface that supports continued solving.}
\end{figure}
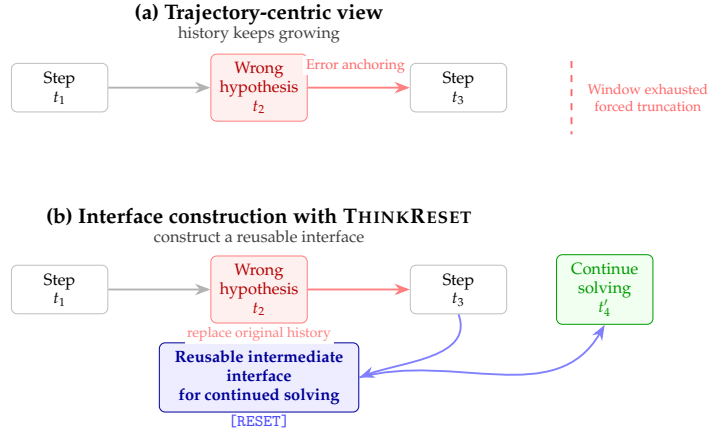

\section{Introduction}
\label{sec:intro}

Large language models have made substantial progress on complex reasoning, from chain-of-thought prompting to recent large reasoning models trained with reinforcement learning \citep{NEURIPS2022_9d560961,Guo_2025,yang2025qwen3technicalreport,kimiteam2025kimik15scalingreinforcement}. Yet their long-horizon performance remains constrained by an underappreciated factor: how intermediate state is managed within the context window. As reasoning traces grow, the model must contend not only with finite context capacity, but also with redundancy accumulation, persistent wrong hypotheses, and the truncation of still-unfinished solution processes; existing work typically addresses these failures through compression, pruning, or test-time control of the reasoning trajectory \citep{xia-etal-2025-tokenskip,pan-etal-2024-llmlingua,hou2026thinkprune,wu2026resumunlockinglonghorizonsearch}.

We argue that this now-common framing is still incomplete. Most prior methods implicitly adopt a trajectory-centric view: reasoning is treated as a growing autoregressive trace, and the main design question becomes how to preserve, compress, or intervene on that trace more effectively. Under bounded context windows, however, the more fundamental question is not whether the trajectory is shorter or better scheduled. It is whether the model can construct a new intermediate interface that can replace discarded history while still supporting continued problem solving.

This leads to our central thesis:

\begin{quote}
    Bounded-context long-horizon reasoning is fundamentally an intermediate interface learning problem: the key is not merely to control or compress the reasoning trajectory, but to construct a reusable intermediate interface that supports continued solving.
\end{quote}

This reformulation shifts the central object from the full trajectory to a reusable intermediate interface, and shifts the optimization target from local fidelity or trajectory control to whether that interface actually supports continued solving.

The same issue appears at the level of training objectives in recent long-chain reasoning systems \citep{Guo_2025,yang2025qwen3technicalreport,kimiteam2025kimik15scalingreinforcement}. In outcome-reward-driven long-chain reinforcement learning, supervision is usually assigned only at the final answer, without explicitly modeling whether the model still preserves the ability to continue solving. When the context window is fixed and the problem is still unfinished as the window fills up, the training signal gradually shifts away from careful reasoning and toward producing an answer as quickly as possible under the remaining context. As a result, the model may abandon detailed reasoning near window exhaustion and jump directly to a guess.

This is not merely a length problem or a generic sparse-reward problem, despite the growing focus on efficient reasoning and overthinking mitigation in recent work \citep{xia-etal-2025-tokenskip,sui2025stop}. It arises because current objectives do not treat a reusable, solvable intermediate interface as an explicit optimization target. As long as the model is trained to care only about the full trajectory or the final answer, it has little incentive to construct a state that can replace history and still support future reasoning. This failure mode therefore exposes a structural mismatch between bounded-context long-horizon reasoning and existing training objectives.

This is also why we do not center the method around local fidelity or length proxies. For long-horizon reasoning, the key issue is not whether the rewritten state can reconstruct the original trace token by token, but whether it can preserve enough problem-solving capability to make progress after reset. We therefore treat the intermediate interface itself as an explicit trainable object and optimize it directly for its ability to support subsequent reasoning.

We therefore propose a learnable rewriting framework in which the intermediate state is not treated as a restatement of history, but as an explicit trainable object. The system can proactively construct such a state at the right time, replace the preceding long trace, and continue reasoning from it. The central question is no longer how to retain as much of the trajectory as possible, but how to learn an interface that is sufficient for continued solving.

To instantiate this framework, we introduce \thinkreset{}. Under a fixed context window, \thinkreset{} triggers interface writeback once context usage reaches a threshold, generates a new intermediate state, and replaces the preceding long trajectory with it. Rather than optimizing local reconstruction, length penalties, or other indirect proxies, we directly optimize post-reset continuation success. The intermediate interface itself is therefore the optimization target.

Accordingly, \thinkreset{} should not be understood merely as a length-control mechanism or a test-time scheduling heuristic. Rather, it is a concrete realization of the shift from trajectory retention to interface construction. Under bounded context windows, gains in long-horizon reasoning come from learning reusable intermediate interfaces, not merely from trimming, retaining, or rescheduling the original trajectory.

\paragraph{Contributions.}
Our main contributions are as follows:
\begin{itemize}
    \item \textbf{Problem reformulation.} We reformulate bounded-context long-horizon reasoning as an \emph{intermediate interface learning problem}. The core challenge is not preserving the full reasoning trace, but learning a reusable interface that can replace history and support continued solving.

    \item \textbf{A unified perspective.} We interpret redundancy accumulation, context overflow, and error anchoring through the lens of missing intermediate interfaces. These are not isolated failures, but different manifestations of the model's inability to construct, update, and reuse effective interfaces.

    \item \textbf{Method instantiation.} We propose \thinkreset{} as a text-space instantiation of this view. Unlike trajectory-retention methods, \thinkreset{} treats reusable intermediate interfaces as explicit, trainable objects and learns them via text-space interface writeback and reset.

    \item \textbf{Training-dynamics analysis.} We identify a key failure mode of outcome-reward-driven long-chain RL under fixed context windows: if the model has not solved the task before the window is exhausted, the final-answer reward encourages it to abandon detailed reasoning and guess. This shows that current objectives do not optimize for a reusable, solvable intermediate interface.

    \item \textbf{Optimization target.} Rather than centering training on trajectory fidelity or length-based proxies, we directly optimize \emph{continuation success} after reset, aligning the objective with interface construction.

    \item \textbf{Empirical validation.} Across multiple long-horizon reasoning benchmarks, learning reusable intermediate interfaces under fixed context windows consistently improves success rate and reasoning stability.
\end{itemize}

\section{Related Work}
\label{sec:related}

\paragraph{From trajectory retention to interface construction.}
We recast bounded-context long-horizon reasoning as an intermediate interface learning problem. From this perspective, most prior work can be understood as pursuing \emph{trajectory retention}: the reasoning trace, whether full or partial, remains the primary object to preserve, compress, schedule, or refresh. In such methods, intermediate states are typically derived from the trajectory through pruning, recoding, scheduling, or external management, rather than learned as explicit interfaces.

By contrast, we focus on \emph{interface construction}. The central object is not the original trace itself, but a trainable intermediate interface whose value lies in replacing history while still supporting subsequent reasoning. This distinction provides a unified way to organize related work.

\paragraph{Trajectory compression and length control.}
Long chain-of-thought reasoning often improves performance on complex tasks, but it also creates context overflow and overthinking issues \citet{NEURIPS2022_9d560961,Guo_2025,yang2025qwen3technicalreport,kimiteam2025kimik15scalingreinforcement,qu2025surveyefficientreasoninglarge,sui2025stop,chen2026thinkdeepjustlong,ma2025reasoningmodelseffectivethinking}. Representative approaches mitigate context limits by pruning, re-encoding, or selectively preserving trajectories, including TokenSkip \citet{xia-etal-2025-tokenskip}, LLMLingua-2 \citet{pan-etal-2024-llmlingua}, step-entropy compression \citet{li2026making}, ThinkPrune \citet{hou2026thinkprune}, R1-Compress \citet{wang2025rcompress}, and Extra-CoT \citet{tang2026efficientlargelanguagereasoning}. In our taxonomy, these are trajectory-retention methods: they ask how to preserve more of the trace under a limited window, whereas \thinkreset{} focuses on interface construction for continued solving.

\paragraph{Latent variables and state representation learning.}
Some work reduces context burden through latent variables or compressed state representations, such as CoLaR \citet{NEURIPS2025_0706261a}, COCONUT \citet{hao2025training}, and LightThinker \citet{zhang-etal-2025-lightthinker}. These methods partially move away from explicit trace retention, but their representations are often less interpretable and are rarely trained as reusable interfaces that directly support future solving. \thinkreset{} instead operates entirely in text space: it introduces neither latent heads nor extra compression units, and it uses the interface itself as the context entry point for subsequent reasoning. Our focus is therefore not on preserving the same process with less explicit history, but on whether the interface can sustain continued solving.

\paragraph{Runtime context refresh and state management.}
In multi-turn interaction and agent systems, converting growing histories into staged states is common; examples include ReSum \citet{wu2026resumunlockinglonghorizonsearch}, Metacognitive Reuse \citet{didolkar2025metacognitivereuseturningrecurring}, and Pensieve / StateLM \citet{liu2026the}. These methods mainly target dialogue history, search, or external memory management. \thinkreset{} is closest in spirit, but narrower in scope: it does not manage dialogue or tool calls, and instead focuses on resets inside a single reasoning trajectory. The key difference is that we model fixed-window reset and reusable interface construction jointly, and directly train the interface through post-reset continuation success.

\paragraph{Recursive frameworks, test-time controllers, and runtime refresh.}
Recent work also studies stronger execution frameworks and runtime-level context management. Recursive reasoning frameworks place subtasks into isolated contexts with explicit call-return structure and can offer formal active-context savings guarantees \citet{yang2026recursivemodelslonghorizonreasoning}. Halo represents a test-time controller that adjusts planning near reasoning boundaries using entropy signals rather than learning the written-back state itself \citet{li2026limitedreasoningspacecage}. COMPASS elevates context management to the agent level through a main agent, a metacognitive planner, and a context manager \citet{wan2025compassenhancingagentlonghorizon}. TIM/TIMRUN performs runtime KV pruning based on task relevance \citet{luo2025contextlimitssubconsciousthreads}. Compared with these approaches, \thinkreset{} is narrower and more direct: it neither modifies the execution engine nor relies on adaptive instability-driven control, but learns reusable intermediate interfaces inside a vanilla autoregressive text trajectory. Our contribution is thus not a stronger theory of context management, but evidence that reusable intermediate interfaces can themselves be trained in vanilla text space and improve post-reset continuation success.

\paragraph{Position of our method.}
Unlike the above lines of work, we do not center the problem on preserving or flexibly scheduling trajectories. We instead ask how to learn an intermediate interface that can replace history and still support continued solving. In that sense, \thinkreset{} is not a simple variant of pruning or control, but a different statement of the core optimization object in bounded-context long-horizon reasoning: a shift from trajectory retention to interface construction.

\section{The \thinkreset{} Method}
\label{sec:method}

\subsection{Mechanism Overview}

We view bounded-context long-horizon reasoning as an \emph{intermediate interface learning} problem. Under a fixed context window $C$, the challenge is not how to retain as much of the trace as possible, but how to construct a reusable interface that still supports solving after the original history is removed. The goal of \thinkreset{} is therefore not prolonged trajectory retention, but a text-space writeback-and-reset mechanism centered on interface construction.

Concretely, once context usage reaches a threshold $\triggerratio C$, the system inserts a trigger prompt $p_{\mathrm{trig}}$ that asks the model to write an intermediate state. Let the conditioning prefix before the first reset be $h_i=(x,r_{1:i},p_{\mathrm{trig}})$, where $x$ is the problem and $r_{1:i}$ is the reasoning trace up to the trigger. The model generates a written-back state $s=\sigma(h_i)$, after which the system replaces the long trace with $x\oplus s$ and continues solving from this new context.

Crucially, $s$ is not treated as a compressed replica of the original trace. It is treated as an explicitly constructed intermediate interface: future reasoning no longer depends on $r_{1:i}$, but only on $x\oplus s$. \thinkreset{} is therefore not a reset heuristic, but an instantiation of reusable intermediate interface learning. Under this view, redundancy accumulation, context overflow, and error anchoring are all consequences of lacking an interface that can replace history and still support continued solving.

If all 8 independent continuations after the first writeback fail, the current interface is still insufficient. The sample is then routed to a later stage in which the model continues reasoning until it again reaches $\triggerratio C$ and performs a second writeback. This allows \thinkreset{} to learn multiple rounds of interface construction under the same fixed context window, rather than relying on a single preserved long trajectory.

In the current version, triggering uses a fixed ratio rather than adaptive semantic boundaries, with $\triggerratio$ treated as a hyperparameter. We also explored semantic triggers, but without robust criteria they induced repeated writebacks and premature resets; see Appendix~\ref{sec:failed-semantic-trigger}. In this paper, the trigger policy is an implementation detail. The core object remains interface construction, not triggering itself.

\subsection{Training Objective}
\label{sec:reward}

If the purpose of the written-back state is to replace history and support continued solving, then the training target should not be local fidelity or a length proxy. It should directly measure how effective that state is as an intermediate interface. We therefore optimize \emph{continuation success}: after history is removed, can the resulting interface still support solving?

Let $\pi_\theta^+$ denote the augmented policy over complete trajectories $\tau$, and let $M(\tau)$ be the peak effective context length during the trajectory; if a reset occurs, the context length is computed after replacement by the written-back state. Over policies satisfying the window constraint, we optimize
\begin{equation}
    \max_{\pi_\theta^+:\,M(\tau)\le C}\;
    \mathbb{E}_{\tau \sim \pi_\theta^+}\left[\Rs(\tau)\right]
    \label{eq:objective}
\end{equation}
where $\Rs$ is not a measure of trajectory fidelity, but of whether the interface supports continued solving.

For a trajectory with one reset, let $h_i=(x,r_{1:i},p_{\mathrm{trig}})$ be the prefix used to generate the writeback, and let $s \sim \pi_\theta(\cdot \mid h_i)$ be the written-back state. We then discard the original history and sample 8 independent continuations from $x\oplus s$, defining
\begin{equation}
    \Rs = \frac{1}{8}\sum_{m=1}^{8}
    \mathbf{1}\!\left[\,\pi_\theta(x \oplus s \leadsto a^\star)_m\,\right]
    \label{eq:reward}
\end{equation}
as the fraction of successful continuations. For the $g$-th sampled writeback trajectory, the sample-level reward is denoted ${\Rs}_g$.

This objective deliberately avoids standard trajectory-retention proxies. In bounded-context long-horizon reasoning, the central question is not whether the written-back state resembles the missing history, but whether it can replace the role previously played by the discarded context in supporting future reasoning. We therefore optimize the continuation success of a reusable intermediate interface directly. The same reward definition is reused for the second reset in Stage 3. We explored alternative objectives based on local fidelity, pre-specified interface completeness, and length terms, but they did not align stably with the actual goal of continued solving; see the Appendix.

\subsection{Three-Stage Training}

\paragraph{Stage 1: cold-start SFT.}
We first collect a small set of manually designed examples (500) whose written-back states still support continued solving after reset, and supervise only the writeback segment after the first reset. This stage is not meant to optimize brevity; it provides a stable initialization for solvable intermediate interfaces and prevents later RL from collapsing into empty writebacks or vague paraphrases under sparse reward.

\paragraph{Stage 2: RLOO after the first reset.}
For each problem $x$, we first sample $G'=16$ trajectories. We retain only those that exceed the threshold $\triggerratio C$ and still end incorrectly; let their number be $G$. These are exactly the samples that expose the core difficulty of bounded-context long-horizon reasoning: the problem is unfinished while the window is nearly exhausted, so a reusable intermediate interface is most needed.

For each retained trajectory, we truncate at $\triggerratio C$, remove the unfinished sentence to avoid semantic discontinuity, and perform the first reset. For trajectory $g$, let $h_{i,g}=(x,r^{(g)}_{1:i},p_{\mathrm{trig}})$ be the prefix before the first reset, and let the policy generate the first writeback state as $s_g \sim \pi_\theta(\cdot \mid h_{i,g})$. We then sample 8 independent continuations from each $s_g$ and use the resulting continuation success ${\Rs}_g$ as reward. If $G<2$, the problem is skipped for the Stage 2 RLOO update.

For the retained samples, RLOO uses the mean reward of the other $G-1$ trajectories in the group as a leave-one-out baseline. Following \citet{ahmadian-etal-2024-back}, the Stage 2 policy-gradient estimator is
\begin{equation}
    \hat{g}_{\mathrm{RLOO}}
    =
    \frac{1}{G}\sum_{g=1}^{G}
    \left(
    {\Rs}_g - \frac{1}{G-1}\sum_{g' \neq g}{\Rs}_{g'}
    \right)
    \nabla_\theta \log \pi_\theta(s_g \mid h_{i,g}),
    \label{eq:rloo}
\end{equation}
where the term in parentheses is the leave-one-out advantage for the $g$-th writeback state. Gradients are applied only to the writeback segment.

This filtering scheme also reveals a central training-dynamics failure mode. Under a fixed context window, outcome-reward-driven long-chain RL does not automatically learn reusable intermediate interfaces. When the problem remains unsolved near window exhaustion, the final-answer reward encourages the policy to stop detailed reasoning and guess. The objective optimizes final correctness, but not the construction of an interface from which solving can continue. Stage 2 addresses this mismatch by explicitly training interface construction on difficult long trajectories and rewarding continuation success rather than terminal outcome alone.

We do not apply this update to two types of trajectories. If a sample finishes correctly before the threshold, interface construction is unnecessary. If a long trajectory exceeds the threshold but still solves the problem correctly, using post-reset continuation success as supervision can encourage the model to stuff nearly complete solutions into the written-back state, degenerating back toward trajectory retention rather than reusable interface learning.

\paragraph{Stage 3: second-reset training on failed samples.}
We collect Stage 2 samples whose 8 post-reset continuations all fail. These are allowed to continue reasoning until they again reach $\triggerratio C$, at which point a second reset is triggered and trained with the same continuation-success reward. Stage 2 begins with up to $G'=16$ raw trajectories per problem and yields at most $G\le16$ retained candidates, corresponding to at most $16\times 8=128$ continuations. Stage 3 is applied only to the subset that completely fails after the first interface.

Mechanistically, Stage 3 is not just an extra reset. It trains hierarchical interface construction: when one intermediate interface is insufficient, the model must write another interface to support even longer-horizon solving. This directly reflects our position that bounded-context long-horizon reasoning is an interface-construction problem, rather than a problem of preserving the full history for as long as possible.

\section{Experiments}
\label{sec:exp}

\subsection{Experimental Setup}

Our goal is not to compare which method shortens the reasoning trace the most. Instead, we test the following claim: under a fixed context window, the key to long-horizon reasoning is whether the model can construct a reusable intermediate interface that still supports solving after history is replaced. Accordingly, our experiments ask two questions: (1) can \thinkreset{}, as a text-space writeback-and-reset framework, improve continuation success under fixed context windows? and (2) do these gains come from \emph{interface construction} rather than generic trajectory retention or test-time resets?

We use Qwen3-8B, Qwen3-14B, and Qwen3-32B as base models, and train on the decontaminated DeepMath-103K dataset \citet{he2025deepmath103klargescalechallengingdecontaminated}. The main paper reports AIME 2024, AIME 2025, ZebraLogic, and AutoLogi, with GPQA-Diamond as a cross-domain benchmark. Together, these tasks cover the typical setting of bounded-context long-horizon reasoning: multi-stage derivation, error accumulation, and solving under a fixed context window.

Our baselines fall into two categories. The first is the \emph{trajectory retention} family, including fixed-ratio trigger + free writeback, length-penalized RLOO, TokenSkip$^\dagger$, and Halo. These represent untrained writeback, RL with a length proxy, trajectory compression, and test-time dynamic intervention, respectively. The second is the \emph{interface construction} family, represented by \thinkreset{}. We are not interested in which method more aggressively preserves, compresses, or refreshes the original trace. We care about which method explicitly treats a solvable intermediate interface as the optimization target. We report Avg@8, the average accuracy over 8 independent continuations, as a direct measure of post-reset continuation success. Detailed training and evaluation settings are given in Appendix~\ref{sec:config-overview}.

\subsection{Main Results}

\begin{table}[http]
    \scriptsize
    \centering
    \caption{Main-task and cross-domain results on the 8B model (Avg@8). We report the mean and 95\% bootstrap confidence interval over 5 random seeds. Improvements over baselines are statistically significant under paired bootstrap tests ($p<0.01$). $\dagger$ indicates additional SFT fine-tuning.}
    \label{tab:main}
    \setlength{\tabcolsep}{3pt}
    \resizebox{\textwidth}{!}{%
        \begin{tabular}{lcccccc}
            \toprule
            \textbf{Method} & \makecell{\textbf{Training}\\\textbf{type}} &
            \textbf{AIME 2024} & \textbf{AIME 2025} & \textbf{ZebraLogic} &
            \textbf{AutoLogi} & \textbf{GPQA-Diamond} \\
            \midrule
            Qwen3-8B             & N/A      & 76.0$\pm$0.8  & 67.3$\pm$0.9 & 84.8$\pm$0.7 & 89.1$\pm$0.6 & 62.0$\pm$1.0 \\
            \midrule
            Fixed-ratio trigger + free writeback & None & 76.6$\pm$0.9  & 68.1$\pm$1.0 & 85.7$\pm$0.8 & 90.1$\pm$0.7 & 62.6$\pm$1.1 \\

            Length-penalized RLOO & RL & 75.3$\pm$1.1  & 67.5$\pm$1.0 & 85.2$\pm$0.9 & 88.3$\pm$0.8 & 61.4$\pm$1.2 \\
            TokenSkip$^\dagger$ & SFT & 72.3$\pm$1.2  & 63.1$\pm$1.3 & 78.4$\pm$1.1 & 82.2$\pm$1.0 & 55.6$\pm$1.3 \\
            Halo & SFT+RL & 79.6$\pm$0.8  & 70.4$\pm$0.9 & 89.2$\pm$0.7 & 93.1$\pm$0.6 & 64.8$\pm$1.0 \\
            \midrule
            \textbf{\thinkreset{}} & SFT+RL &
            \textbf{81.3$\pm$0.6} & \textbf{73.2$\pm$0.7} &
            \textbf{91.5$\pm$0.5} & \textbf{93.5$\pm$0.5} &
            \textbf{66.7$\pm$0.8} \\
            \bottomrule
        \end{tabular}%
    }
\end{table}

Table~\ref{tab:main}, together with the 14B and 32B results in Tables~\ref{tab:main14B} and~\ref{tab:main32B}, shows a consistent pattern. Under fixed context-window constraints, the gains do not come from more aggressive preservation, compression, or scheduling of the original trajectory; they come from learning an intermediate interface that can replace history and still support continued solving. \thinkreset{} achieves the best or tied-best performance across model sizes and tasks, suggesting that this formulation is empirically robust.

The comparison with fixed-ratio trigger + free writeback is particularly important. Both methods share the same trigger rule and context-window constraint; the difference is not whether a reset occurs, but whether the written-back state is explicitly trained to serve as a reusable intermediate interface. The consistent gain therefore cannot be attributed to context refresh itself. It reflects better interface construction. In other words, \thinkreset{} does not optimize the refresh action; it optimizes the effectiveness of the intermediate interface as a new entry point for continued solving.

The comparison with length-penalized RLOO shows that simply pushing the model toward shorter traces does not solve the core difficulty of bounded-context long-horizon reasoning. If the objective is centered on a length proxy, the model is more likely to learn to expand less rather than to preserve the ability to keep solving after history is replaced. \thinkreset{} improves performance because it directly optimizes continuation success, making the intermediate interface, rather than trajectory length, the core training target.

The comparison with TokenSkip and other heavy compression baselines supports the same distinction. Trajectory-retention methods mainly answer the question, ``How can we preserve more of the existing trace under a bounded window?'' \thinkreset{} instead answers, ``How can we construct an intermediate interface that can replace history?'' The former may improve window utilization but need not improve post-replacement solvability; the latter directly optimizes that ability.

Halo shows that stronger test-time intervention can help, but still does not substitute for explicitly learning intermediate interfaces. Test-time controllers mainly help decide when to intervene, whereas \thinkreset{} additionally learns what must be written back so that solving can continue. This shifts inference-time state management from trajectory scheduling toward interface construction.

Overall, the main results suggest that under bounded context windows, the key optimization object is not trajectory retention, but interface construction. As a text-space instantiation of this perspective, \thinkreset{} consistently improves success rate and robustness in long-horizon reasoning.

\subsection{Ablation Study}
\label{sec:ablation}

\begin{table}[http]
    \small
    \centering
    \caption{Ablation results on the 8B model (Avg@8). Each row removes one design component.}
    \label{tab:ablation}
    \resizebox{\textwidth}{!}{%
        \begin{tabular}{lcccc}
            \toprule
            \textbf{Variant} & \textbf{AIME 2024} & \textbf{AIME 2025} & \textbf{ZebraLogic} & \textbf{AutoLogi} \\
            \midrule
            \thinkreset{} (full) & \textbf{81.3$\pm$0.6} & \textbf{73.2$\pm$0.7} & \textbf{91.5$\pm$0.5} & \textbf{93.5$\pm$0.5} \\


            $-$ RL writeback optimization (SFT only)  & 77.4$\pm$1.0 & 68.5$\pm$1.1 & 86.1$\pm$0.9 & 90.3$\pm$0.8 \\
            $-$ second reset stage & 80.9$\pm$0.7 & 72.7$\pm$0.8 & 90.6$\pm$0.6 & 93.0$\pm$0.6 \\
            $-$ easy-sample filtering & 78.3$\pm$0.9 & 70.6$\pm$1.0 & 88.4$\pm$0.8 & 91.2$\pm$0.7 \\
            $-$ cold-start SFT (direct RLOO) & 77.9$\pm$1.1 & 70.1$\pm$1.0 & 87.7$\pm$0.9 & 90.3$\pm$0.8 \\
            $-$ cold-start SFT ($+$ use summary prompt instead) & 76.4$\pm$0.6  & 67.8$\pm$0.8 & 85.5$\pm$0.7 & 89.7$\pm$0.5  \\
            \bottomrule
        \end{tabular}
    }
\end{table}

The ablations show that the gains of \thinkreset{} do not come from any isolated trick, but from a training design centered on reusable intermediate interface learning.

Removing RL writeback optimization causes a large drop, showing that a small amount of SFT alone is insufficient to reliably learn states that support future solving. This matches our training-dynamics analysis: without explicitly optimizing continuation success, the model lacks pressure to construct truly reusable interfaces and instead tends to produce superficial paraphrases or local restatements that do not reliably support continuation.

Removing easy-sample filtering also hurts performance. The reason is not merely reduced data size; it is target dilution. Training with a summary-oriented objective leads to worse final performance, suggesting that the summary objective is not well aligned with continuation success.

Removing the second reset stage also degrades performance, indicating that some hard instances require hierarchical interface construction rather than a single writeback. This supports the mechanism view of \thinkreset{}: it is not a one-off refresh heuristic, but a framework that allows the model to learn multiple interfaces within a single long-horizon reasoning process.

Finally, removing cold-start SFT also hurts performance, showing that direct RLOO is unstable under sparse and delayed continuation-success rewards.

\subsection{Sensitivity to Interface-Construction Timing}
\label{sec:sensitivity}

\begin{table}[http]
    \small
    \centering
    \caption{Sensitivity to the fixed-ratio trigger threshold $\triggerratio$ on AIME 2025 (8B model, Avg@8). All settings share the same training configuration. $\triggerratio=3/4$ is the most stable choice.}
    \label{tab:threshold-sensitivity}
    \begin{tabular}{lccccc}
        \toprule
        \textbf{Metric} & $1/2$ & $2/3$ & $\mathbf{3/4}$ & $4/5$ & $5/6$ \\
        \midrule
        \textbf{AIME 2025} & 72.1$\pm$0.4 & 72.9$\pm$0.5 & \textbf{73.2$\pm$0.7} & 72.6$\pm$0.5 & 71.9$\pm$0.4 \\
        \bottomrule
    \end{tabular}
\end{table}

This experiment examines the stability of interface construction under different trigger times. If triggered too early, the model is forced to write back before accumulating sufficient problem structure; if triggered too late, the prefix is richer but too little context remains after reset.
The results suggest that $\triggerratio=3/4$ provides the most stable trade-off.

\paragraph{Qualitative observation.}
Manual inspection of 50 long-chain logic problems shows that when the trigger comes too late, the model is more likely to hit window exhaustion before the writeback is fully formed.

\section{Limitations}

Our evidence is concentrated on bounded-context long-horizon reasoning in math and logic; broader validation on scientific reasoning, program reasoning, and agentic multi-step tasks remains limited. In addition, \thinkreset{} still relies on a small amount of cold-start SFT to stabilize interface learning. Finally, the current implementation uses fixed-ratio triggering and at most two resets, so more flexible triggering and deeper interface hierarchies remain for future work.

\section{Conclusion}
\label{sec:conclusion}

We introduced \thinkreset{} and framed bounded-context long-horizon reasoning as an \emph{intermediate interface learning problem}. Our central claim is that, under a fixed context window, the key challenge is not merely to preserve, prune, or schedule an existing trajectory, but to construct a reusable intermediate interface that can replace history and still support continued solving. Through this reformulation, we further identified a key failure mode of outcome-reward-driven long-chain RL: when the model has not finished solving before the window is exhausted, training encourages premature guessing rather than continued careful reasoning, because current objectives do not optimize reusable solvable interfaces.

As a text-space instantiation of this perspective, \thinkreset{} explicitly constructs and learns reusable interfaces through interface writeback and reset, and directly optimizes post-reset continuation success. Across multiple bounded-context long-horizon reasoning benchmarks, this approach consistently improves success rate and reasoning stability under a fixed context window.

Overall, our results support a simple conclusion: under bounded context windows, the key optimization object is not trajectory retention, but interface construction. We hope this perspective provides a more unified starting point for training-objective design, inference-time state management, and reusable intermediate interface modeling in long-horizon reasoning.

\section*{Reproducibility Statement}
Code and materials will be released.

\section*{Ethics Statement}
No new privacy or copyright risks beyond standard language-model training and evaluation.

\appendix
\section{Appendix Guide}

The appendix is organized as follows. We first clarify how benchmark evidence is organized; then we report additional 14B and 32B results on the same evaluations as in the main paper; next we summarize several failed attempts that were not retained in the main paper, including semantic-stage triggering, fixed structured templates, CRP, reward proxies based on writeback length or continuation length, forward-continuation probes, and frozen external judges; finally, we provide a qualitative case analysis.

\section{Benchmark Organization}

To reduce ambiguity in benchmark selection, we make the evidential organization explicit.

\paragraph{Main-paper evidence.}
The main paper uses \textbf{AIME 2024, AIME 2025, ZebraLogic, and AutoLogi} as the primary evidence on difficult, long-chain, staged reasoning, and adds GPQA-Diamond as a cross-domain check. Together, these benchmarks support our main claim: under a fixed context window, the core challenge is not simply retaining more history, but training a reusable intermediate interface that can replace history and support continued solving.

\paragraph{Supplementary evidence.}
The appendix further reports results from 14B and 32B models on the same evaluations. This organization keeps the main paper focused on the central mechanism evidence while preserving enough cross-model support for close inspection.

\section{Additional Results for Qwen3-14B}

\begin{table}[H]
\small\centering
\caption{Results of the 14B model on the same evaluation suite as the main paper (Avg@8). We report the mean and 95\% bootstrap confidence interval over 5 random seeds. Improvements over baselines are statistically significant under paired bootstrap tests ($p<0.01$). The format follows Table~\ref{tab:main}.}
\resizebox{\textwidth}{!}{%
\begin{tabular}{lccccc}
\toprule
\textbf{} & \textbf{AIME 2024} & \textbf{AIME 2025} & \textbf{ZebraLogic} & \textbf{AutoLogi} & \textbf{GPQA-Diamond} \\
\midrule
Qwen3-14B & 79.3$\pm$0.9 & 70.4$\pm$1.0 & 88.5$\pm$0.8 & 89.2$\pm$0.7 & 64.0$\pm$1.1 \\
\midrule
Fixed-ratio trigger + free writeback & 81.9$\pm$0.8 & 72.3$\pm$0.9 & 91.5$\pm$0.7 & 92.1$\pm$0.7 & 66.2$\pm$1.0 \\
Length-penalized RLOO & 77.1$\pm$1.0 & 69.3$\pm$1.1 & 88.8$\pm$0.9 & 88.2$\pm$0.8 & 63.3$\pm$1.2 \\
TokenSkip$^\dagger$ & 74.3$\pm$1.2 & 65.5$\pm$1.3 & 81.1$\pm$1.0 & 82.4$\pm$0.9 & 58.2$\pm$1.3 \\
Halo & 81.3$\pm$0.8 & 73.1$\pm$0.9 & 91.9$\pm$0.7 & 93.5$\pm$0.6 & 66.2$\pm$1.0 \\
\midrule
\textbf{\thinkreset{}} & \textbf{83.9$\pm$0.6} & \textbf{75.8$\pm$0.7} & \textbf{95.2$\pm$0.5} & \textbf{93.7$\pm$0.5} & \textbf{68.9$\pm$0.8} \\
\bottomrule
\end{tabular}
}

\label{tab:main14B}
\end{table}

\section{Additional Results for Qwen3-32B}

\begin{table}[H]
\small\centering
\caption{Results of the 32B model on the same evaluation suite as the main paper (Avg@8). We report the mean and 95\% bootstrap confidence interval over 5 random seeds. Improvements over baselines are statistically significant under paired bootstrap tests ($p<0.01$). The format follows Table~\ref{tab:main}.}
\resizebox{\textwidth}{!}{%
\begin{tabular}{lccccc}
\toprule
\textbf{} & \textbf{AIME 2024} & \textbf{AIME 2025} & \textbf{ZebraLogic} & \textbf{AutoLogi} & \textbf{GPQA-Diamond} \\
\midrule
Qwen3-32B & 81.4$\pm$0.8 & 72.9$\pm$0.9 & 88.8$\pm$0.7 & 87.3$\pm$0.8 & 68.4$\pm$1.0 \\
\midrule
Fixed-ratio trigger + free writeback & 83.8$\pm$0.7 & 74.4$\pm$0.8 & 91.6$\pm$0.6 & 91.2$\pm$0.7 & 69.3$\pm$0.9 \\
Length-penalized RLOO & 78.9$\pm$0.9 & 71.2$\pm$1.0 & 89.0$\pm$0.8 & 87.3$\pm$0.9 & 66.2$\pm$1.1 \\
TokenSkip$^\dagger$ & 76.4$\pm$1.1 & 67.3$\pm$1.2 & 81.4$\pm$1.0 & 81.1$\pm$1.0 & 61.5$\pm$1.3 \\
Halo & 83.1$\pm$0.7 & 75.2$\pm$0.8 & 92.2$\pm$0.6 & 91.9$\pm$0.7 & 70.3$\pm$0.9 \\
\midrule
\textbf{\thinkreset{}} & \textbf{86.1$\pm$0.5} & \textbf{77.3$\pm$0.6} & \textbf{95.4$\pm$0.5} & \textbf{92.9$\pm$0.6} & \textbf{72.6$\pm$0.8} \\
\bottomrule
\end{tabular}
}

\label{tab:main32B}
\end{table}

\section{Failed Attempt: Triggering on Natural Semantic Stages}
\label{sec:failed-semantic-trigger}

We initially explored a more human-like trigger strategy: instead of using a fixed window-ratio threshold, we performed writeback and reset when the model appeared to finish a local subproblem and transition into a new semantic stage.

The intuition behind this idea is clear. If the trigger lands exactly on a stage boundary, the written-back state is more likely to correspond to a semantically closed intermediate state. We therefore tried several simple semantic heuristics, such as summary-like phrases, discourse shifts, or explicit local conclusions. In practice, however, the approach did not yield stable gains and exposed four problems.

\paragraph{Stage boundaries are hard to define robustly.}
In long-chain logical reasoning, stages are rarely as clean as paragraph boundaries. The model often revisits the same local conclusion with different wording, or appears to transition before the subproblem is truly closed. Trigger rules then misinterpret superficial discourse changes as genuine stage boundaries.

\paragraph{It induces repeated writebacks and premature resets.}
If writeback is triggered before the current state has closed semantically, the model may repeatedly write back and re-expand the same local issue. The resulting state may look structurally complete while still corresponding to an unresolved fragment, which hurts continuation stability.

\paragraph{Stage-wise interface construction itself creates structural failure modes.}
If each semantic stage is written back separately, information that was once necessary can later become obsolete, yet stage-wise writeback cannot easily remove stale information from earlier states. If the system instead reconstructs an interface from the full prefix each time, it repeatedly revisits content that has already been processed. Either way, simple semantic staging does not reliably yield interfaces that support future solving.

\paragraph{Training signals become harder to attribute.}
When trigger locations are chosen by unstable semantic heuristics, it becomes difficult to disentangle whether a success or failure comes from interface quality or from a poor trigger time. This coupling increases variance and complicates interpretation.

\section{Failed Attempt: Defining the Interface with a Fixed Structured Template}
\label{sec:failed-fixed-template}

We also explored fixing the written-back state to an explicit slot-based template, for example separating confirmed conclusions, eliminated paths, current direction, and key constraints.

This design improves readability and local inspection, but under our end-to-end continuation-success objective it has two drawbacks.

\paragraph{Templates pre-specify the form of the interface.}
The information that must be preserved at reset time does not always fit naturally into the same fixed slots. Once the structure is fixed in advance, the model tends to spend capacity on filling the template rather than learning what kind of intermediate interface best supports future solving.

\paragraph{Templates pull the paper back toward surface form.}
If the main paper centers on a fixed template, reviewers are likely to read the work as a question about which writeback template is better, rather than whether the post-reset interface truly improves end-to-end continuation success. That weakens the main thesis that bounded-context long-horizon reasoning is an interface-construction problem.

\paragraph{Design implication.}
We therefore do not place fixed structured templates at the center of the current version and retain them only as an early exploratory interface form.

\section{Failed Attempt: Using CRP as a Local Interface Probe}
\label{sec:failed-crp}

We also tried a constraint-reconstruction probe (CRP) to directly measure whether the written-back state preserves key intermediate conclusions. The idea is to extract local conclusions from the pre-reset trajectory and test whether they can be reconstructed from the written-back state alone.

This design is locally computable and does provide a cheap approximate signal, but we ultimately did not keep it as a central objective.

\paragraph{CRP optimizes local reconstructability, not continuation success.}
A state that fails to reconstruct some local conclusion may still support successful solving through a different path. Conversely, a state that passes a local reconstruction test may still fail to support the final solve. CRP is therefore not equivalent to our actual target: whether the interface supports continued problem solving.

\paragraph{CRP pulls the paper back toward local fidelity.}
If CRP were central, the discussion would shift toward whether the probe is reasonable and whether the extracted local conclusions are general enough, rather than whether the written-back state constitutes a reusable intermediate interface. That would turn the paper back into an analysis of local interface quality rather than a mechanism paper about bounded-context long-horizon reasoning.

\paragraph{Design implication.}
We therefore do not use CRP as the main training objective and retain it only as an exploratory local probe.

\section{Failed Attempt: Using Writeback Length as a Reward Proxy}
\label{sec:failed-short-summary-reward}

We also tried directly adding a length reward on the written-back state, giving higher reward to shorter writebacks in the hope of encouraging more concise intermediate states.

The intuition is straightforward: a shorter written-back state seems to occupy less context. In practice, however, it quickly led to reward hacking.

These attempts were motivated by efficiency-style proxy objectives, but they failed precisely because they do not optimize reusable intermediate interfaces.

\paragraph{The model learns omission, not interface construction.}
When shortness is rewarded directly, the easiest strategy is not to preserve the right information more efficiently, but to omit key constraints, skip eliminated paths, or replace necessary state with vague generic statements. The written-back text becomes shorter, but continuation success often collapses.

\paragraph{Length rewards optimize surface form, not interface function.}
A shorter written-back state does not necessarily mean a better interface; it may simply contain less information. Such rewards conflate being shorter with being functionally stronger, and the model optimizes the easier surface property instead of the harder but more important interface-construction ability.

\paragraph{Design implication.}
We therefore do not reward writeback length directly. Instead, we optimize whether the written-back state supports future solving.

\section{Failed Attempt: Using Post-Writeback Continuation Length as a Reward Proxy}
\label{sec:failed-short-continuation-reward}

We also tried using the length of the continuation after writeback as a reward proxy, under the assumption that shorter continuations indicate a better written-back state.

This seems more appealing than rewarding shorter writebacks because it attempts to align the reward with downstream effects. In practice, however, it introduces a strong boundary bias.

\paragraph{The model shifts more solving into the written-back state.}
Once shorter post-writeback continuations are rewarded, the model's easiest strategy is not to learn better abstractions, but to push more reasoning into the written-back state itself. This can shorten the continuation while making the written-back state increasingly bloated and less like an intermediate interface.

\paragraph{The reward optimizes boundary migration, not interface quality.}
A shorter continuation does not necessarily imply a better interface; it may only indicate that solving was moved earlier. Such a reward systematically favors the partition ``longer writeback + shorter continuation'' rather than preserving exactly the information needed for future solving.

\paragraph{Design implication.}
We therefore do not use post-writeback continuation length as a reward term.

\section{Failed Attempt: Forward-Continuation Consistency as an Interface Probe}
\label{sec:failed-forward-consistency}

We also explored a probe that appears closer to practical needs: instead of reconstructing past conclusions, generate future reasoning from both the full context and the written-back state, and compare whether the two continuations agree at key decision points.

The intuition is sensible. If the written-back state is truly a substitute for the full context, then future reasoning from it should roughly match future reasoning from the full context. In practice, however, the probe ran into two fundamental issues.

\paragraph{It wrongly penalizes equivalent but different trajectories.}
Many math problems do not have a unique future reasoning path. Even if two states both support correct solving, the model may choose different decompositions, validation orders, or intermediate expressions. Whether one compares next-step KL divergence or agreement on the next key value, superficial path differences are easily mistaken for poor interface quality.

\paragraph{Forward sampling itself is high variance.}
Unlike backward reconstruction, forward-continuation consistency compounds three factors at once: interface quality, sampling randomness, and path diversity. The resulting signal is noisy and hard to attribute, while also requiring a task-dependent definition of what counts as a key decision point.

\paragraph{Design implication.}
We therefore do not adopt forward-continuation consistency as the main reward. It is closer to rewarding whether the model continues along the original path than whether the written-back state forms a reusable interface for continued solving.

\section{Failed Attempt: A Frozen External Model as Interface Judge}
\label{sec:failed-frozen-judge}

To avoid circularity in training, we also tried introducing a frozen external model as a judge of intermediate-interface quality. The judge performs the probe under the written-back state, so the same model no longer both writes the state and grades it.

This setup is cleaner on paper, but it introduces new biases and instability.

\paragraph{Interface quality is outsourced to another model's preferences.}
The judge's capability, reasoning style, and inductive biases directly shape the reward. If the judge is weaker, the signal is capped by its competence. If its style differs from that of the trained model, optimization can drift from preserving the right interface toward satisfying the judge's preferences.

\paragraph{It can become a new form of judge overfitting.}
Once the frozen judge becomes a stable reward source, the model may learn to produce states that are easy for that judge to approve rather than genuinely useful for future solving.

\paragraph{Design implication.}
We therefore do not use a frozen external judge. Although it reduces circularity, it turns interface evaluation into a cross-model alignment problem and simultaneously increases complexity, compute, and sources of bias.

\section{Training and Evaluation Configuration Overview}
\label{sec:config-overview}

For easy reference, we summarize the key experimental settings here. We use Qwen3-8B, Qwen3-14B, and Qwen3-32B as base models. Training follows a three-stage pipeline: cold-start SFT, RLOO after the first reset, and a second-reset stage on failed samples. The main reward is the average success rate over 8 continuations after the first or second reset; the grouped-sampling details are described in Section~\ref{sec:reward}. Main-table results use Temperature=0.6, TopP=0.95, TopK=20, and MinP=0. We compare methods mainly on Avg@8 over long-chain logic tasks and GPQA-Diamond. All experiments use a $32$k context window, i.e., $C=32$k.

\paragraph{Form of the writeback segment and trigger prompt.}
The writeback segment $s$ is the text generated immediately after the trigger prompt $p_{\mathrm{trig}}$; we do not introduce separate structured slots or external state objects. In implementation, it is simply a natural-language intermediate interface written by the model. The segment begins immediately after the trigger prompt and ends when the model stops the segment or when the segment reaches the context limit and is truncated. The current trigger template can be written as: ``Based on the above reasoning, write back an intermediate state that supports continued solving, and keep only the information necessary for future progress.'' The same template is used for \thinkreset{} and the free-writeback baseline.
The prompt asks the model to construct a state that preserves only the information necessary to continue solving, rather than to summarize the preceding trajectory.

\paragraph{Context window and number of resets.}
All tasks, models, and baselines use the same context window $C=32$k. Stage 2 allows at most one reset. Stage 3 allows a second reset only when all 8 continuations after the first reset fail. Therefore, the number of resets is mechanically restricted to $\{0,1,2\}$ in the current version, and all appendix analyses follow this setting. At evaluation time, each trajectory can use at most two resets.

\paragraph{Compute and comparison fairness.}
All methods use the same decoding settings and the same Avg@8 evaluation protocol. For \thinkreset{}, training rewards also use the average over 8 continuations, keeping the training objective aligned with evaluation. Length-penalized RLOO and TokenSkip are trained for the same GPU time budget as \thinkreset{}.

\paragraph{Free-writeback baseline.}
The fixed-ratio trigger + free-writeback baseline shares exactly the same trigger prompt $p_{\mathrm{trig}}$, writeback stopping rule, context-window constraint, and decoding parameters as \thinkreset{}. The only difference is that its written-back state $s$ is not optimized by reset-conditioned RL; it is simply a natural-language continuation generated at the same trigger point. This isolates the gain from learning the interface itself, rather than confounding it with different prompts, stopping rules, or window settings.

\paragraph{Halo baseline.}
Our Halo implementation follows the test-time controller described in the original paper \citet{li2026limitedreasoningspacecage}. It uses the same backbone model, monitors layer-averaged attention entropy during reasoning, and accumulates an instability score with a threshold controller. Once the score exceeds a tolerance threshold, Halo triggers a semantic rewrite and trajectory re-initialization, then continues reasoning from the new state representation. The original paper implements Halo in vLLM and uses staged rewriting plus history reset as a form of state correction. Its appendix reports stable performance around a tolerance threshold of 5 and entropy sensitivity around 0.85. Our Halo baseline keeps this controller structure and the same fixed-window setup, and serves to compare entropy-driven test-time control against interface construction through trainable text-space interfaces.

\section{Qualitative Case Analysis}

We conclude with a qualitative comparison on a long-chain logic problem. Under the same fixed context-window constraint, \thinkreset{} constructs two intermediate interfaces and reaches the same final answer as a full CoT baseline with a larger context window. In both cases, the trigger occurs when context usage is already close to the threshold and the problem remains unsolved. At that point, \thinkreset{} does not simply retain more history; it rewrites the preceding process into a reusable intermediate interface and continues from it. As a result, solving proceeds even though the original history is no longer present, instead of being cut off at the context limit.

The key point in this case is not that one reset happened to work. It is that the intermediate interface is sufficient to carry the remaining solution process. The role of \thinkreset{} is therefore not to patch a trajectory locally, but to explicitly construct an interface that can replace history and support future reasoning in bounded-context long-horizon reasoning.

\section{Why \thinkreset{} Is Not Reducible to Learned Summarization}
\label{sec:not-summarization}

A natural question is whether the writeback in \thinkreset{} is essentially just learned summarization under a different terminology. This section argues that the two are structurally distinct, along four dimensions: optimization target, behavior on erroneous trajectories, output content space, systematic patterns in failed experiments, and empirical evidence from the free-writeback baseline.

\paragraph{The optimization targets are fundamentally different.}
The training signal of learned summarization is almost always some form of fidelity to the source: ROUGE, reconstruction loss, semantic similarity, KL divergence, or other metrics that measure how well the compressed text preserves information from the original. Even when reinforcement learning is used, the reward typically remains anchored to faithfulness to the source text. In contrast, the reward in \thinkreset{} is continuation success: the original reasoning trajectory has been entirely discarded, and the sole criterion is whether solving can continue correctly from the written-back state within the remaining context window. This is not a variant of fidelity; it is a functional criterion. In short, learned summarization asks ``how much of the past is preserved,'' while interface construction asks ``how much of the future is supported.'' These two optimization targets diverge---and can even oppose each other---precisely when the original trajectory contains errors, redundancy, or suboptimal decompositions.

\paragraph{Behavior on erroneous trajectories is opposite.}
This is the most discriminative difference. If the preceding reasoning chain contains wrong hypotheses, dead ends, or redundant paths, a faithful summary will compress these errors into the output---because its objective is to reproduce the content of the source. An interface optimized for continuation success, however, has an incentive to actively discard erroneous paths, correct wrong conclusions, and reorganize the problem state. That is, on erroneous trajectories, a good summary and a good interface behave in opposite directions: the summary preserves errors, while the interface corrects them. The error-anchoring phenomenon in Figure~1 is a concrete manifestation of this distinction: trajectory-centric summarization compresses and retains error anchoring, whereas interface construction can break it. If the writeback were merely a summary, error anchoring would not be alleviated by a reset; the empirical gains of \thinkreset{} indicate that the writeback learns behavior that goes beyond faithful summarization.

\paragraph{The content space of the writeback strictly contains that of a summary.}
Summarization carries an implicit constraint: the output content should originate from the input content, i.e., a source--target correspondence exists. An interface has no such constraint. A good writeback may re-derive an intermediate conclusion via a path different from the original trajectory, reorganize the representation of constraints, or even introduce a new decomposition strategy that never appeared explicitly in the original trace---as long as doing so supports subsequent solving. Formally, if a summary is viewed as a function of the input trajectory, then the content space of an interface strictly contains that of a summary: summaries are constrained to be faithful compressions of the original trajectory, while interfaces are constrained only by whether they support future solving. This means an interface can introduce reformulations, corrections, and reorganizations that have no direct antecedent in the original trace.

\paragraph{Six failed experiments trace a spectrum from summarization to interface construction.}
The six failed attempts in the appendix are not isolated engineering notes; they systematically reveal why summarization-style objectives are insufficient for bounded-context long-horizon reasoning. They can be organized as a spectrum from fidelity to functionality:

\begin{itemize}
    \item The constraint-reconstruction probe (CRP; Appendix~\ref{sec:failed-crp}) is essentially a test of local summarization fidelity---whether local conclusions from the original trajectory can be reconstructed from the writeback. It failed because reconstructing local conclusions does not entail supporting future solving.
    \item The writeback-length reward (Appendix~\ref{sec:failed-short-summary-reward}) optimizes the surface form of the summary---shorter is better. It led to reward hacking: the model learned to omit critical information rather than to preserve information more efficiently.
    \item The post-writeback continuation-length reward (Appendix~\ref{sec:failed-short-continuation-reward}) attempted to measure summary quality indirectly through downstream effects---shorter continuations should indicate better summaries. It led to boundary migration: the model shifted solving into the writeback rather than learning a better interface.
    \item Forward-continuation consistency (Appendix~\ref{sec:failed-forward-consistency}) tests whether the writeback can replace the original text and produce the same subsequent behavior---the ultimate fidelity criterion for summarization. It failed because it penalized equivalent but different solving paths.
    \item The frozen external judge (Appendix~\ref{sec:failed-frozen-judge}) outsources writeback quality assessment to another model---essentially delegating the summarization fidelity judgment. It devolved into a cross-model alignment problem.
    \item The fixed structured template (Appendix~\ref{sec:failed-fixed-template}) pre-specifies the slot structure of the writeback---a prior constraint on the form of the summary. It limited the model's freedom to learn which interface best supports future solving.
\end{itemize}

The common pattern across these failures is that every proxy based on information fidelity, local reconstructability, or surface form failed to produce stable improvements; only directly optimizing continuation success worked. This systematic pattern is itself empirical evidence that interface $\neq$ summary.

\paragraph{Summary.}
Across these four dimensions, we argue that the writeback in \thinkreset{} cannot be reduced to learned summarization. The core difference lies not in implementation details but in the fundamental orientation of the optimization target: summarization aims at faithful compression of the past, while interface construction aims at supporting the solving of the future. In bounded-context long-horizon reasoning, this orientation difference leads to systematically different behavior in error handling, content generation, and training-signal design. Both the experimental results and the systematic failure patterns of fidelity-based proxies support this conclusion.

\end{document}